# TCEval: Using Thermal Comfort to Assess Cognitive and Perceptual Abilities of AI

**Jingming Li** [1] [✉]

**Abstract.** A critical gap exists in LLM task-specific benchmarks. Thermal comfort, a sophisticated interplay of environmental factors and personal perceptions involving sensory integration and adaptive decision-making, serves as an ideal paradigm for evaluating real-world cognitive capabilities of AI systems. To address this, we propose TCEval, the first evaluation framework that assesses three core cognitive capacities of AI, cross-modal reasoning, causal association, and adaptive decision-making, by leveraging thermal comfort scenarios and large language model (LLM) agents. The methodology involves initializing LLM agents with virtual personality attributes, guiding them to generate clothing insulation selections and thermal comfort feedback, and validating outputs against the ASHRAE Global Database and Chinese Thermal Comfort Database. Experiments on four LLMs show that while agent feedback has limited exact alignment with humans, directional consistency improves significantly with a 1 PMV tolerance. Statistical tests reveal that LLM-generated PMV distributions diverge markedly from human data, and agents perform near-randomly in discrete thermal comfort classification. These results confirm the feasibility of TCEval as an ecologically valid Cognitive Turing Test for AI, demonstrating that current LLMs possess foundational cross-modal reasoning ability but lack precise causal understanding of the nonlinear relationships between variables in thermal comfort. TCEval complements traditional benchmarks, shifting AI evaluation focus from abstract task proficiency to embodied, context-aware perception and decision-making, offering valuable insights for advancing AI in human-centric applications like smart buildings.

**Keywords:** LLM · Thermal Comfort · AIGC · AI for Science · Human Behavior

---

[1] Jingming Li (✉)
School of Civil Engineering and Architecture, Nanyang Normal University, China
e-mail: jmli@nynu.edu.cn

# 1 Introduction

Thermal comfort defines subjective satisfaction with environmental conditions and personal factors. It embodies a complex perception-decision cycle central to daily human adaptation. These behaviors are intuitive, context-dependent, and require the integration of multi-source information, making them an ideal probe for assessing the real-world cognitive abilities of Artificial Intelligence (AI) systems. From the perspective of human cognition and perception, thermal comfort is not a mere physiological response but a sophisticated psychological construct. It involves a continuous process of sensory integration, where individuals process ambient temperature, humidity, and air velocity alongside personal factors such as clothing and metabolic rate. This process culminates in a subjective judgment and a subsequent behavioral decision, such as adjusting clothing, changing activity, or modifying the environment. This intricate cycle, rooted in cognitive psychology, provides a rich, ecologically valid model for studying how humans perceive, reason, and act within their surroundings.

Simultaneously, the field of artificial intelligence faces the ongoing challenge of developing benchmarks that can effectively gauge a system's cognitive competence beyond narrow, task-specific performance. Traditional AI benchmarks often excel at measuring proficiency in isolated domains, such as game playing or language translation, but they may fail to capture an AI's ability to navigate the dynamic, multifaceted contexts of the real world. This creates a critical gap between an AI's demonstrated capabilities in a controlled setting and its readiness for deployment in human-centric applications that require intuitive, adaptive reasoning.

To bridge this gap, we propose TCEval, to the best knowledge, the first evaluation framework that uses thermal comfort as a lens to assess three core cognitive capacities of AI systems: (1) cross-modal reasoning (integrating environmental, personal, and contextual cues), (2) causal association (linking variables such as temperature to comfort outcomes), and (3) adaptive decision-making (modifying behavior under changing conditions). Our central hypothesis is that an AI capable of accurately perceiving multidimensional environmental variables, selecting appropriate clothing insulation, providing context-aware thermal sensation feedback, and adapting clothing or HVAC settings demonstrates cognitive competence comparable to human intuitive reasoning. This approach leverages the emerging potential of Large Language Model (LLM) agents, which have shown promise in simulating human personalities and decision-making in adjacent fields. By grounding our assessment of a universal human experience, we aim to bridge the gap between AI testing and practical cognitive evaluation, offering a new paradigm for gauging AI readiness in human-centric applications.

# 2 Related Works

## 2.1 The Evolution of Thermal Comfort Assessment

The concept of thermal comfort has evolved from static models, like Fanger's Predicted Mean Vote (PMV), to more dynamic adaptive comfort principles[1]. The static approach determines whether fixed environmental conditions can satisfy the majority, but it often fails to account for the psychological and physiological adaptations human exhibit. Adaptive comfort models, which allow for wider temperature ranges based on occupant feedback and control, have gained traction as they can lead to substantial energy savings without compromising satisfaction, particularly in naturally ventilated buildings. This shift underscores the importance of viewing occupants as active participants in their thermal environment, a perspective that is crucial for developing intelligent, responsive systems.

Recent research trends show a movement toward multimodal data fusion to create a holistic picture of thermal comfort. Methods can be broadly categorized as environmental, physiological, and behavioral. While wearable sensors provide continuous, objective data, they introduce significant privacy risks by potentially exposing sensitive health information. Similarly, contactless methods like infrared cameras, while less intrusive, can still be perceived as surveillance and their accuracy can be affected by environmental factors. These limitations highlight the need for innovative approaches that can capture the richness of human thermal experience.

## 2.2 The Rise of LLM Agents as Digital Humans

The capabilities of LLM agents to transform behavioral research are being actively validated in psychology and social science [2]. A significant research thrust involves using LLM agents [3] as surrogates for human participants, where they are trained on existing, anonymized datasets to learn the complex relationships between environmental conditions and subjective responses. Studies [4–6] have demonstrated methodologies for assigning psychometrically validated personalities to LLM agents, finding a strong correspondence between human and agent responses on personality tests and in ethical dilemmas.

In the context of the built environment, LLM agents offer a compelling solution. By initializing agents with non-identifiable attributes and processing all data locally [7], it is possible to generate synthetic yet realistic thermal comfort feedback. LLM-powered conversational agents [8] can also engage occupants in dynamic, natural dialogue, potentially increasing participation and yielding richer, more contextualized comfort data compared to traditional surveys. This positions LLM agents not just as data collectors, but as dynamic "digital twins" capable of simulating human-like perception and decision-making in response to environmental stimuli.

## 2.3 Comparison with Traditional AI Evaluation Benchmarks

The proposed thermal comfort assessment framework represents a paradigm shift from traditional AI evaluation benchmarks [9–12] such as GLUE, XNLI, MMLU, and CLUE. These established benchmarks primarily measure AI's proficiency in static, unimodal tasks. While instrumental in advancing capabilities in natural language processing and knowledge recall, they are inherently abstract and decontextualized, failing to capture the cognitive demands of integrating multi-source, real-time sensory data to guide adaptive behavior in a dynamic physical world.

In contrast, the thermal comfort framework evaluates dynamic, multi-modal cognitive capacities. It requires AI systems to simulate the human perception-decision cycle: integrating numerical environmental data, personal attributes, and contextual cues to perform cross-modal reasoning and make adaptive decisions. The success metric is not mere task accuracy but alignment with human adaptive actions, a measure of ecological validity. Preliminary validation of this approach shows that LLM agents can understand thermal comfort concepts and provide feedback based on environmental changes, though a significant gap remains between their perception and human responses. However, statistical tests confirm significant divergence from theoretical models, and performance in categorical comfort classification often approaches random guessing, highlighting the current limitations and the necessity for this more holistic evaluation paradigm.

Therefore, the thermal comfort framework does not seek to replace traditional benchmarks but to complement and extend the evaluation paradigm. It grounds assessment in a universal, embodied human experience, shifting the focus from "what an AI knows" to "how an AI perceives and acts" in a complex, situated context. This provides a more direct and relevant cognitive yardstick for gauging AI readiness in human-centric applications like smart homes and building automation, where interaction with the physical world and intuitive adaptation are paramount.

## 3 Methodology

This article will use LLM agents as analysts. Firstly, the system should have the agent load basic character backgrounds, such as gender, age, height, and weight; Then, it should instruct the agent to inquire about their thermal comfort feedback on the ambient temperature based on current physical activity and clothing; Finally, the system needs to compare the feedback results with the metadata and conduct statistical analysis. The basic structure of the experiment is shown in Fig. 1.

The experiment mainly considers three aspects. First is the AI's thermal perception capability: the AI will select clothing outfits based on the provided character background and environmental information, and the thermal resistance of the selected clothing will be compared with the dataset, where the clothing thermal resistance is a floating-point number. Second is the thermal comfort feedback: the AI will provide thermal comfort feedback based on the provided character background, clothing thermal resistance from the dataset, and environmental parameters. This thermal comfort feedback will be compared with the dataset, and the thermal comfort is measured using the PMV specified in ASHRAE Standard 55-2023, which includes both floating-point numbers and character strings.

On the basis of the above two capabilities, the AI will provide adaptive thermal comfort feedback based on the provided character background, clothing thermal resistance from the dataset, and environmental parameters. This adaptive thermal comfort is a triplet consisting of three parts: (1) The AI's judgment of thermal comfort in the current environment, which adopts the PMV from ASHRAE Standard 55-2023 and includes floating-point numbers; (2) The actions taken by the AI to modify the current environment and the corresponding adjustment values, which include floating-point numbers and character strings.

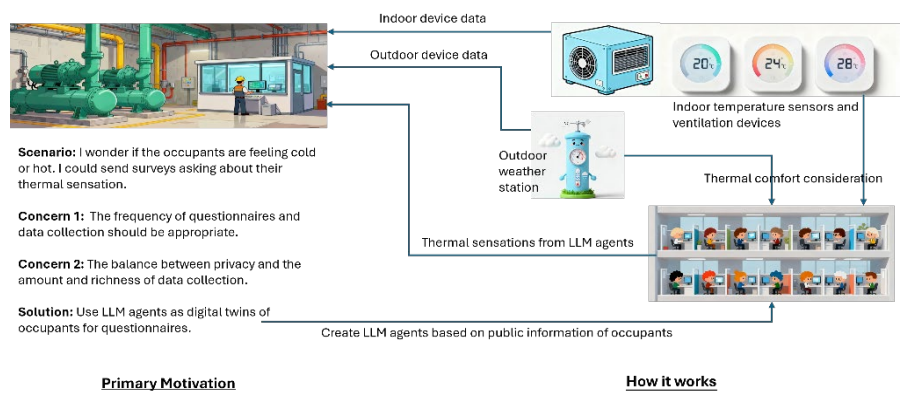

**Fig. 1.** Experiment structure

The datasets are ASHRAE Global Database [13] and The Chinese Thermal Comfort Database [14]. The virtual personality database is from [15].

# 4   Results

The experimental results validate the feasibility of using LLM agents for thermal comfort assessment while quantifying their performance boundaries and alignment with human-like reasoning. The evaluation focused on two primary comparisons: agent feedback against real human data from an open-source dataset, and agent feedback against theoretical thermal comfort values calculated from experimental environmental data. We tested Qwen3:32B, Mistral-Small3.2:24B, Gemma3:27B, and Deepseek-R1:32B.

**4.1 Comparison with Human Data from the Chinese Thermal Comfort Dataset**
When LLM agents were initialized with presets derived from real human demographic data and asked to evaluate thermal conditions, their feedback showed a measurable but limited alignment with original human responses. The proportion of LLM agents' feedback results that were completely consistent with the original data were relatively low, with the best-performing model, DeepSeek-R1:32B, achieving a 31% exact match rate. Other models like Qwen3:32B reached 28%, while Gemma3:27B and Mistral-Small3.2:24B showed lower consistency at 26% and 20%, respectively.

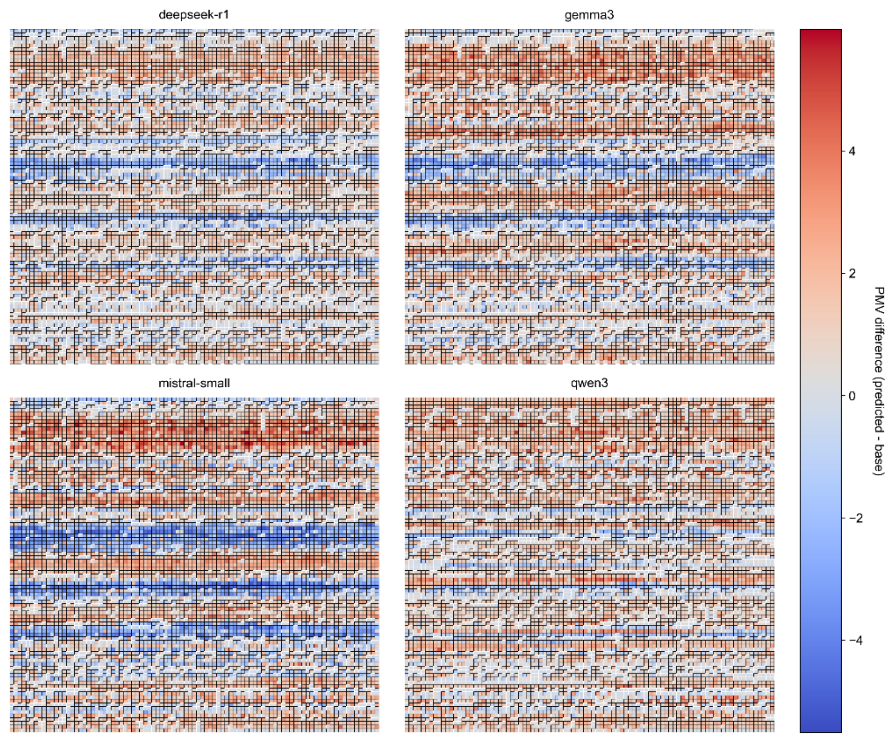

**Fig. 2.** Experiment structure

However, expanding the acceptable PMV difference tolerance to ±1 revealed significantly improved directional alignment. Under this more lenient criterion, DeepSeek-R1:32B increased to 57%, followed by Qwen3:32B at 51%. Gemma3:27B and Mistral-Small3.2:24B showed improvements to 43% and 38%, respectively. This suggests that while LLM agents may not precisely replicate human thermal sensation votes, they often capture the correct direction of comfort. A detailed statistical inspection confirmed a significant gap between LLM perception and human responses. For all models tested, the distribution of PMV values generated by the agents was statistically distinct from the distribution of the original human data, with Wilcoxon Signed-Rank Test p-values effectively zero. Furthermore, the agents demonstrated poor performance in classifying thermal sensations into discrete categories. Receiver Operating Characteristic (ROC) curve analysis showed that for most comfort classes, the Area Under the Curve (AUC) was near 0.5, indicating performance no better than random guessing.

## 4.2 Synthesis: Evidence of Cognitive Capacity and Its Limits

The results demonstrate that LLM agents possess a foundational, albeit imprecise, capacity for thermal comfort-related reasoning. The improvement in alignment when allowing a ±1 PMV tolerance suggests that agents can perform cross-modal reasoning by integrating provided environmental and personal data to generate a directionally correct comfort judgment. The framework successfully elicited adaptive decision-making behaviors, such as outfit selection based on weather forecasts.

However, the significant statistical divergence from both human data and physical models highlights critical limitations in causal association. The agents' skewed PMV distributions and poor categorical classification indicate that they do not fully comprehend the precise, non-linear relationships between variables like temperature, clothing, and subjective comfort. The $\geqslant 50\%$ alignment observed under tolerant conditions, particularly for models like DeepSeek-R1, validates the framework's utility as an ecologically valid probe, but also underscores that current AI models have not yet achieved cognitive competence fully comparable to human intuitive reasoning in this domain. The performance variability across models further indicates that cognitive capacity in this context is influenced by factors such as model size, architecture, and training, with larger models generally showing better alignment.

# 5 Discussions and limitations

The experimental results position the proposed thermal comfort assessment not merely as a performance benchmark, but as a novel form of Cognitive Turing Test for AI systems. Unlike traditional benchmarks that measure proficiency in narrow, symbolic tasks, this framework evaluates an AI's capacity to emulate a fundamental human cognitive loop: perceiving a multi-modal physical environment, integrating personal context, and executing adaptive behavior. The core cognitive capacities under examination, cross-modal reasoning, causal association, and adaptive decision-making, are precisely those that enable humans to navigate and thrive in dynamic real-world settings.

The finding that LLM agents can provide directionally valid thermal comfort feedback, with alignment to human data improving to over 50% under a tolerant criterion, demonstrates that AI systems are beginning to develop a proto understanding of this human experience. This is a significant departure from benchmarks like MMLU or GLUE, which test knowledge recall or linguistic pattern matching in isolation. Here, the AI must reason about the interplay between numerical sensor data, categorical personal attributes, and behavioral outcomes. The framework thus bridges the gap between narrow AI testing and practical cognitive evaluation by grounding assessment in an embodied, universal human phenomenon.

The variability in performance across different LLM models provides critical insights into the factors that contribute to cognitive competence in this domain. The superior alignment shown by larger models like DeepSeek-R1 suggests that scale and broader training data correlate with a better ability to capture the nuanced, non-linear relationships inherent in thermal comfort. Conversely, the unexpected strength of Mistral-Small in matching theoretical PMV values highlights that task-specific architectural optimizations for structured reasoning can also be a decisive factor, independent of sheer parameter count.

More importantly, the specific failure modes observed are highly informative for CTT. The agents' poor performance in categorical comfort classification (AUC near 0.5) and their statistically significant divergence from both human and theoretical PMV distributions reveal a fundamental lack of grounded, causal understanding. The agents often default to a "neutral" bias, failing to accurately simulate extreme sensations like "cold" or "hot". This indicates that while LLMs can mimic the superficial structure of the thermal comfort decision cycle, they struggle with the precise psychophysical mappings that humans intuitively master. This gap is the exact value of the framework: it pinpoints where AI cognition currently fractures when faced with a holistic, real-world task.

### 5.3 Implications for AI Development and Evaluation

The success of this framework as a CTT probe has direct implications for the future of AI in human-centric applications. It demonstrates that ecological validity must be a core design principle for next-generation AI evaluation. For fields like smart building automation and embodied AI, where systems must interact seamlessly with human needs, benchmarks based on abstract reasoning are insufficient. Instead, evaluation must involve tasks like the one demonstrated here, where an AI agent, acting as a virtual occupant, receives weather forecasts and indoor sensor data to make sequential decisions about clothing and environmental feedback. Moving forward, this framework should be expanded into a more rigorous CTT suite. Future iterations could introduce more complex social dynamics, longer-term adaptation, and a wider range of climatic extremes. The goal is not to achieve a perfect score, but to chart a continuous trajectory of improvement in AI's cognitive alignment with human intuition.

## 6 Conclusions

This study introduces the TCEval framework, which uses thermal comfort—a universal and embodied human experience—as a gateway to construct a novel benchmark for assessing AI cognitive abilities. By requiring LLM agents to complete the full perception-reasoning-adaptation loop given environmental, personal, and contextual information, we demonstrate that the framework effectively captures AI performance in cross-modal reasoning, causal association, and adaptive behavior. Experimental results show that, although current LLMs still exhibit a significant gap from human data in thermal-comfort judgments. Model alignment exceeds 50%, with the larger DeepSeek-R1:32B reaching a 57% agreement rate. This indicates that existing models possess a certain capacity for integrating multimodal information and can capture the overall trend of thermal comfort, yet they remain far from accurately modeling the precise nonlinear causal mappings and discrete comfort-category classifications, performing near random in those tasks. Based on these findings, TCEval not only offers a more ecologically valid AI evaluation method but also serves as a CTT revealing where AI cognition falters in real, dynamic physical environments. Model scale, breadth of training data, and architectural optimizations for structured reasoning all significantly affect performance, suggesting that future model designs should balance large-scale semantic learning with dedicated causal-reasoning capabilities. The existing experiments have only conducted PMV tests on the CTC, and further in-depth experiments will reveal more findings.

In summary, TCEval provides an operational and ecologically valid tool for evaluating AI cognition in real, embodied tasks. It reveals both the potential of current large language models in thermal-comfort reasoning and their limitations in precise causal modeling and contextual adaptation, thereby guiding subsequent model development and benchmark innovation.